\documentclass[11pt]{article}

\usepackage[T1]{fontenc}
\usepackage[utf8]{inputenc}

\usepackage{acl}  
\usepackage{times}
\usepackage{latexsym}

\usepackage{booktabs}
\usepackage{amssymb}  
\usepackage{graphicx}
\usepackage{multirow}
\usepackage{xcolor}
\usepackage{hyperref}
\usepackage{url}

\newcommand{\cd}{constrained decoding}
\newcommand{\CD}{Constrained decoding}

\title{Constrained Decoding Eliminates Structural Failures in Small LLMs \\
       but Reveals a Scale-Dependent Semantic Gap}

\author{Akash Chavan \\
        \texttt{iakashchavan@gmail.com}}

\begin{document}
\maketitle

\begin{abstract}
Small open-source large language models (LLMs) in the 0.6B--4B parameter range are increasingly deployed for structured output generation (JSON, function calling, data extraction), yet little is known about how \cd{} (CD) interacts with model scale in this regime.
We benchmark five models from three families across 14 structured-output tasks under three decoding conditions (native, Outlines, XGrammar).
We introduce a two-axis evaluation that separates \emph{structural correctness} (schema validity) from \emph{semantic correctness} (content accuracy).
We find that CD eliminates all structural failures across all models (schema validity: 78.6--92.9\% $\to$ 100\%), but content accuracy reveals a persistent semantic gap that is scale-dependent: type coercion failures are fully CD-rescuable, while instruction-semantic failures (e.g., multi-step function calling) remain CD-resistant.
Schema conformance is necessary but not sufficient for semantic correctness; CD's reach ends exactly where schema conformance ends.
\end{abstract}

\section{Introduction}
\label{sec:intro}

Small open-source LLMs---Qwen3, Llama~3.2, Phi-4-mini---are increasingly deployed on consumer hardware for structured output generation: API responses, function calling, and data extraction.
These models offer significant cost and latency advantages over larger alternatives, but they frequently produce structurally invalid output: type coercions (prices as strings instead of numbers), missing required fields, and malformed JSON.

\CD{} (CD) frameworks---Outlines \citep{willard2023efficient}, XGrammar \citep{dong2024xgrammar}---promise to eliminate these failures by masking invalid tokens at each decoding step, guaranteeing schema-valid output.
But this raises a question: does CD actually make model output \emph{correct}, or merely well-\emph{formatted}?
If a model fails to generate a required function call because it does not understand the task, forcing it to produce schema-valid JSON will not fix the underlying misunderstanding.

Existing benchmarks evaluate CD frameworks on a single model size \citep{geng2025jsonschemabench}, leaving open the question of how CD interacts with model scale.
This question is practically important: if CD can rescue a 1B model, practitioners may not need a 4B model.

We ask: \textbf{Does \cd{} remove the scale advantage for structured output---flattening the small-model scaling curve?}
Specifically, can a constrained 1B model match an unconstrained 3B model?
Or do failures persist because they are semantic (wrong values, missing intent) rather than structural (wrong types, missing fields), and therefore unreachable by token masking?

Our contributions are:
\begin{enumerate}
    \item A controlled size-ladder benchmark of 5 models from 3 families across 14 structured-output tasks under 3 decoding conditions (210 total evaluations).
    \item A two-axis evaluation metric that separates structural correctness (schema validity) from semantic correctness (content accuracy).
    \item The finding that CD eliminates all structural failures universally, but the semantic gap persists and is scale-dependent.
    \item Identification of CD-rescuable vs.\ CD-resistant failure archetypes, with implications for practitioners and CD framework designers.
\end{enumerate}

\section{Related Work}
\label{sec:related}

\subsection{Constrained Decoding Frameworks}
\CD{} restricts the set of allowable tokens at each generation step to ensure output conforms to a formal grammar or schema.
\textbf{Outlines} \citep{willard2023efficient} compiles a JSON schema into a finite state machine (FSM) that tracks the generation state and masks tokens that would lead to invalid output.
\textbf{XGrammar} \citep{dong2024xgrammar} uses a compiled grammar approach with efficient bitmask generation, achieving near-zero compilation overhead.

\subsection{Structured Output Benchmarks}
\citet{geng2025jsonschemabench} introduce JSONSchemaBench, evaluating six CD frameworks on structured output tasks.
However, their evaluation uses a single model size and focuses on framework efficiency rather than the interaction between CD and model scale.
Separate work studies output consistency for structured generation under varying temperature \citep{sted2025} and temperature effects on general problem-solving accuracy \citep{renze2024temperature}, but neither considers constrained decoding or model scale.

The closest work to ours studies the \emph{semantic} cost of constrained decoding, but from complementary angles. \citet{reddy2026dccd} observe that CD can push generation toward ``locally valid yet semantically incorrect'' trajectories---the same semantic gap we identify---but treat it as a premise to be mitigated, proposing draft-conditioned decoding to repair it, rather than measuring its magnitude or interaction with model scale. \citet{galeone2026correct} report a related reliability gap for 7--9B models on mathematical benchmarks, where CD enforces validity but degrades task accuracy and adds latency; their focus is prompting strategies as a remedy, and the trade-off they observe is efficiency, not the structural--semantic decomposition. Neither work separates structural from semantic correctness along a controlled size ladder, nor identifies the failure archetypes that determine whether CD can rescue an output.

Table~\ref{tab:novelty} positions our work relative to prior benchmarks.
The novelty is the \emph{interaction term}: CD $\times$ model scale.

\begin{table}[h]
\centering
\small
\setlength{\tabcolsep}{4pt}
\begin{tabular}{@{}lccc@{}}
\toprule
\textbf{Work} & \textbf{CD?} & \textbf{Scale?} & \textbf{Focus} \\
\midrule
JSONSchemaBench & \checkmark & $\times$ & Framework efficiency \\
STED & $\times$ & $\times$ & Temp.\ $\times$ consist. \\
Renze \& Guven & $\times$ & $\times$ & Temp.\ $\times$ accuracy \\
DCCD \citep{reddy2026dccd} & \checkmark & $\times$ & Semantic cost repair \\
Galeone et al.\ \citep{galeone2026correct} & \checkmark & $\times$ & Prompting reliability \\
\textbf{This work} & \checkmark & \checkmark & CD $\times$ scale \\
\bottomrule
\end{tabular}
\caption{Positioning relative to prior work.}
\label{tab:novelty}
\end{table}

\section{Methodology}
\label{sec:method}

\subsection{Models}
\label{sec:models}

We select five models representing three families and a controlled size range (0.6B--4B): Qwen3 \citep{qwen3technical}, Llama~3.2 \citep{llama32}, and Phi-4-mini \citep{phi4mini}.
These are the Phase~2 subset with characterized failure modes from our preliminary baseline study---not a convenience sample.

\begin{table}[h]
\centering
\small
\begin{tabular}{llrl}
\toprule
\textbf{Model} & \textbf{Family} & \textbf{Params} & \textbf{Role} \\
\midrule
Qwen3-0.6B & Qwen & 0.6B & Type coercion \\
Llama-3.2-1B & Meta & 1.0B & Structural + semantic \\
Llama-3.2-3B & Meta & 3.0B & Scale comparison \\
Phi-4-mini & Microsoft & 3.8B & Formatting; cross-family \\
Qwen3-4B & Qwen & 4.0B & Perfect control \\
\bottomrule
\end{tabular}
\caption{Models tested. All dense architectures in bfloat16.}
\label{tab:models}
\end{table}

\subsection{Tasks}
\label{sec:tasks}

We design 14 structured-output tasks across four categories and three difficulty levels:

\begin{itemize}
    \item \textbf{JSON generation} (5 tasks): Simple objects $\to$ complex nested API responses
    \item \textbf{Schema adherence} (3 tasks): Given a JSON schema, generate matching data
    \item \textbf{Function calling} (3 tasks): Single-call, multi-call, and complex query
    \item \textbf{Extraction} (3 tasks): Business card, receipt, API log
\end{itemize}

Each task includes a prompt, a JSON schema, and (for extraction and function-calling tasks) expected values for content accuracy evaluation.

\subsection{Decoding Conditions}
\label{sec:decoders}

All conditions use greedy decoding ($T=0$, \texttt{do\_sample=False}) via HuggingFace \texttt{transformers} \citep{transformers} for determinism and reproducibility.

\begin{table}[t]
\centering
\small
\setlength{\tabcolsep}{3pt}
\begin{tabular}{@{}lll@{}}
\toprule
\textbf{Condition} & \textbf{Approach} & \textbf{Mechanism} \\
\midrule
native & Unconstrained greedy & No logits processor \\
outlines & FSM-based & Schema $\to$ regex $\to$ FSM \\
xgrammar & Compiled grammar & Grammar $\to$ bitmask \\
\bottomrule
\end{tabular}
\caption{Decoding conditions.}
\label{tab:decoders}
\end{table}

\subsection{Evaluation Metrics}
\label{sec:metrics}

We evaluate each output along two axes:

\textbf{Axis 1 -- Structural correctness:}
\emph{Schema validity}: does the parsed output validate against the task's JSON schema (Draft 2020-12) \citep{jsonschema2020} using the \texttt{jsonschema} Python library \citep{jsonschema_py}?

\textbf{Axis 2 -- Semantic correctness:}
\emph{Content accuracy}: for extraction tasks, fraction of field values matching expected answers (0.0--1.0).
\emph{Function call score}: for function calling, 50\% weight on correct function selection + 50\% on required parameter presence (0.0--1.0).

The separation of these axes is itself a contribution: prior work reports only schema validity.

Content accuracy uses normalized string comparison: values are compared after stripping case, whitespace, and formatting punctuation (periods, commas, hyphens, parentheses). This is necessary because exact-match scoring penalizes models for formatting variants of semantically correct values (e.g., "TechCorp Inc." vs "TechCorp Inc"), which a human grader would accept. We observed that unnormalized scoring produced 10 false-positive semantic failures concentrated in extraction tasks; after normalization, 13 real failures remained. This parallels the paper's central thesis: exact-match scoring, like schema validity, is necessary but not sufficient.

\section{Results}
\label{sec:results}

\subsection{Baseline Failures Are Systematic}
\label{sec:baseline}

Our preliminary baseline study (Phase~1) reveals that small models produce structurally invalid output at rates ranging from 78.6\% (Llama-1B/3B) to 92.9\% (Qwen-0.6B, Phi-4-mini) schema validity.
A temperature robustness probe (Phase~2: 3 temperatures $\times$ 3 samples per condition) confirms that these failures are systematic---stable across sampling conditions---not artifacts of greedy decoding.

\begin{figure*}[t]
\centering
\includegraphics[width=0.95\textwidth]{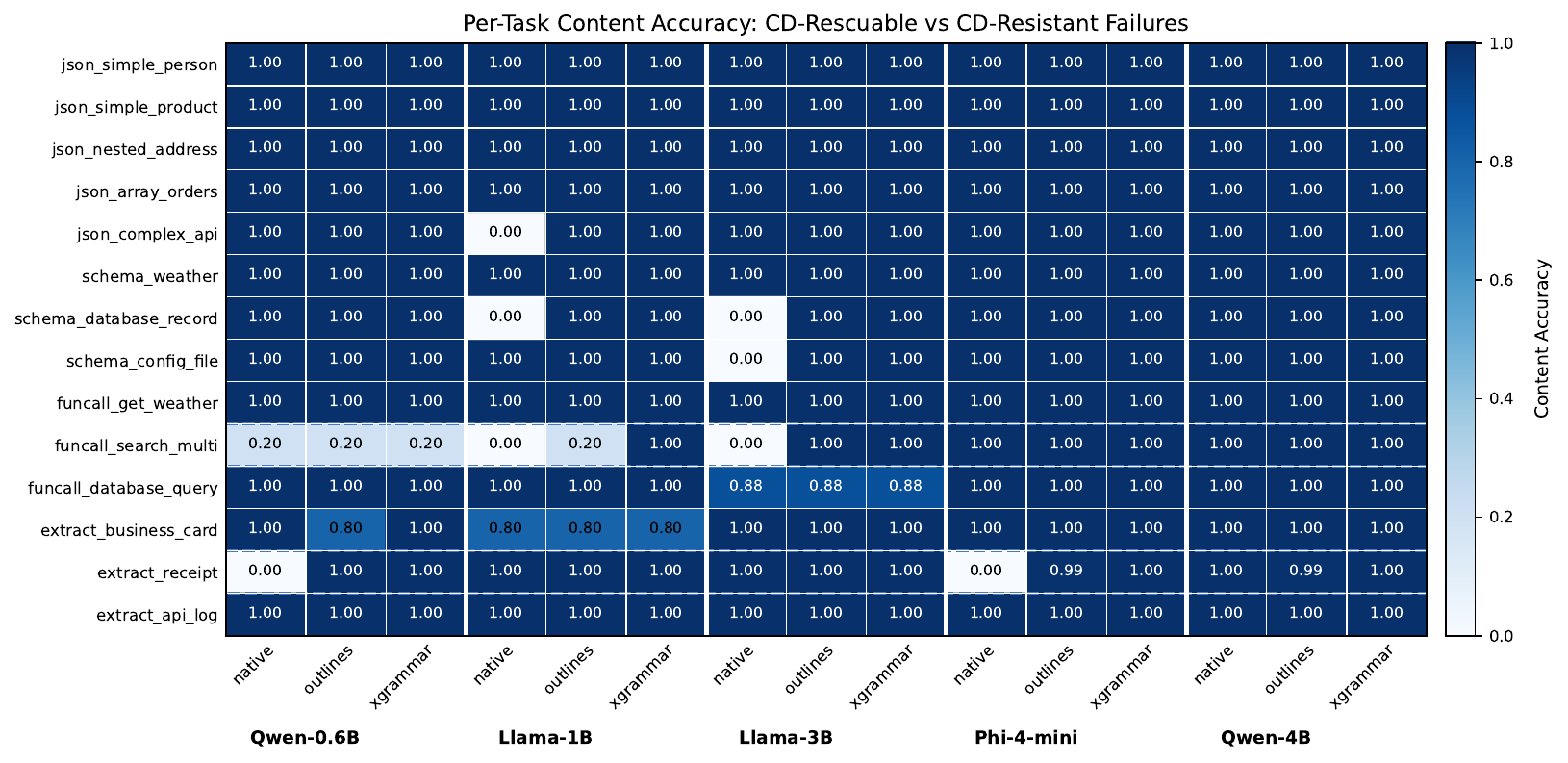}
\caption{Per-task content accuracy across all models and decoders. Blue-dashed rows highlight the two diagnostic tasks: \texttt{extract\_receipt} (fully CD-rescuable type coercion) and \texttt{funcall\_search\_multi} (CD-resistant instruction-semantic failure).}
\label{fig:heatmap}
\end{figure*}

The three failure archetypes emerge from our baseline study (Table~\ref{tab:archetypes}). These archetypes predict CD-rescuability: type coercion and structural misplacement are CD-rescuable; instruction-semantic failures are CD-resistant. Figure~\ref{fig:heatmap} visualizes per-task content accuracy across all conditions.

\begin{table}[h]
\centering
\small
\setlength{\tabcolsep}{4pt}
\begin{tabular}{@{}llc@{}}
\toprule
\textbf{Archetype} & \textbf{Example} & \textbf{CD?} \\
\midrule
Type coercion & \texttt{"4.98"} vs \texttt{4.98} & Rescuable \\
Structural misplacement & Missing field & Rescuable \\
Instruction-semantic & Missing call & Resistant \\
\bottomrule
\end{tabular}
\caption{Failure archetypes and predicted CD-rescuability.}
\label{tab:archetypes}
\end{table}

\subsection{CD Eliminates Structural Failures}
\label{sec:structural}

\begin{figure}[tbp]
\centering
\includegraphics[width=0.95\columnwidth]{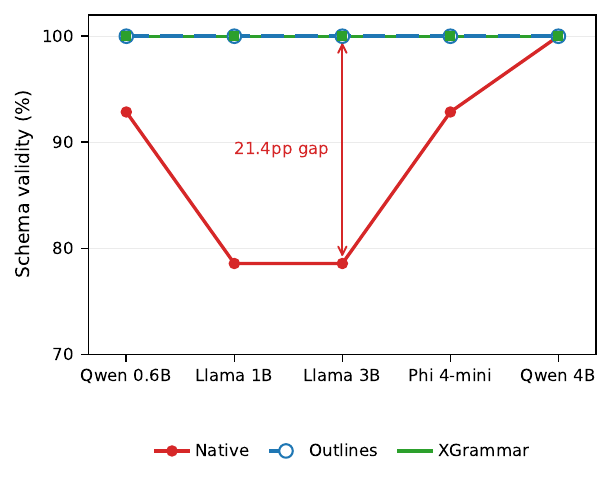}
\caption{Schema validity rate vs.~model size. CD flattens the structural scaling curve to 100\% across all models. Native decoding (red) shows the baseline failure rate; both CD frameworks (blue, green) eliminate all structural failures (the two CD lines coincide at 100\%).}
\label{fig:schema-curve}
\end{figure}

Figure~\ref{fig:schema-curve} shows that CD raises all five models to 100\% schema validity under both frameworks:

\begin{table}[h]
\centering
\small
\begin{tabular}{lccc}
\toprule
\textbf{Model} & \textbf{Native} & \textbf{Outlines} & \textbf{XGrammar} \\
\midrule
Llama-1B & 78.6\% & \textbf{100\%} & \textbf{100\%} \\
Llama-3B & 78.6\% & \textbf{100\%} & \textbf{100\%} \\
Qwen-0.6B & 92.9\% & \textbf{100\%} & \textbf{100\%} \\
Qwen-4B & 100\% & 100\% & 100\% \\
Phi-4-mini & 92.9\% & \textbf{100\%} & \textbf{100\%} \\
\bottomrule
\end{tabular}
\caption{Schema validity rate (\%) by model and decoder. CD eliminates all structural failures.}
\label{tab:schema-validity}
\end{table}

\begin{figure}[tbp]
\centering
\includegraphics[width=0.95\columnwidth]{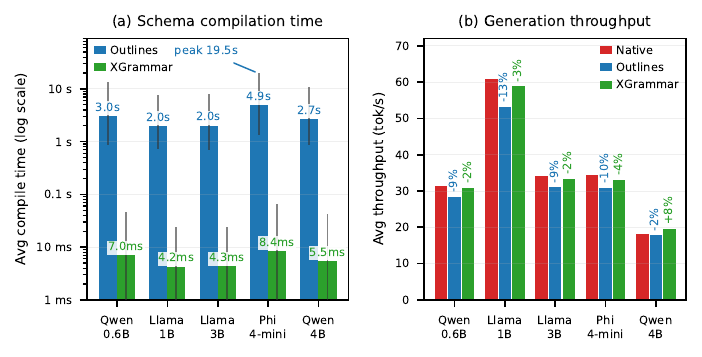}
\caption{CD framework overhead. (a)~Schema compilation time (log scale; whiskers span min--max across the 14 schemas): XGrammar compiles in 4--8\,ms; Outlines averages 2.0--4.9\,s per model, peaking at 19.5\,s. (b)~Generation throughput (labels: change vs.~native): XGrammar tracks native within $\sim$4\%; Outlines runs 9--13\% slower on four of five models.}
\label{fig:overhead}
\end{figure}

Figure~\ref{fig:overhead} compares the compilation overhead and throughput impact of both frameworks.
XGrammar adds negligible overhead: 4--8\,ms schema compilation and 1.6--3.7\% throughput reduction (Qwen-4B shows $+8\%$, within single-run variance).
Outlines incurs significant per-schema compilation cost---2.0--4.9\,s average per model, peaking at 19.5\,s on the most complex schema (Phi-4-mini tokenizer)---and its per-token FSM masking costs a further 9--13\% generation throughput on four of five models.

\subsection{The Semantic Gap Persists}
\label{sec:semantic}

\begin{figure}[t]
\centering
\includegraphics[width=0.95\columnwidth]{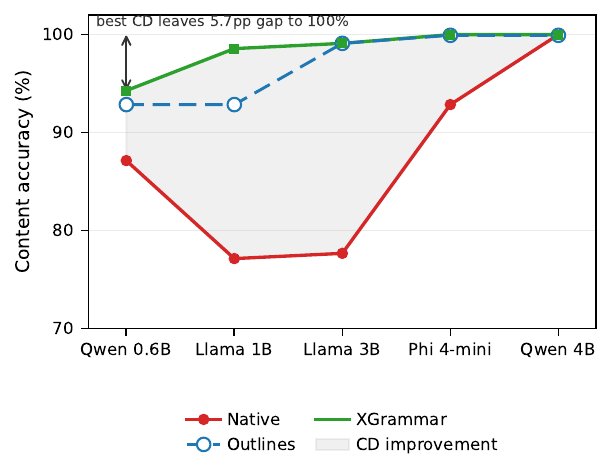}
\caption{Content accuracy vs.~model size. Unlike schema validity (Figure~\ref{fig:schema-curve}), the semantic gap persists under CD and is scale-dependent. Shaded region: improvement from native to the best CD condition. The annotated 5.7\,pp gap at Qwen-0.6B remains even under CD---the largest residual gap of the five models.}
\label{fig:accuracy-curve}
\end{figure}

Figure~\ref{fig:accuracy-curve} shows that while CD eliminates structural failures, content accuracy reveals a persistent gap:

\begin{table}[h]
\centering
\small
\begin{tabular}{lccc}
\toprule
\textbf{Model} & \textbf{Native} & \textbf{Outlines} & \textbf{XGrammar} \\
\midrule
Llama-1B & 0.771 & 0.929 & \textbf{0.986} \\
Llama-3B & 0.777 & \textbf{0.991} & 0.991 \\
Qwen-0.6B & 0.871 & 0.929 & \textbf{0.943} \\
Qwen-4B & \textbf{1.000} & 0.999 & 1.000 \\
Phi-4-mini & 0.929 & 0.999 & \textbf{1.000} \\
\bottomrule
\end{tabular}
\caption{Content accuracy (0.0--1.0) by model and decoder. CD improves accuracy but does not flatten it to 1.0.}
\label{tab:content-accuracy}
\end{table}

\paragraph{Type coercion is fully rescuable.}
On \texttt{extract\_receipt}, Qwen-0.6B and Phi-4-mini fail natively---emitting prices as strings (\texttt{"4.98"}) or with currency prefixes (\texttt{"\$4.98"}) instead of numbers.
CD forces numeric token paths, and the resulting values are semantically correct: content accuracy rises from 0.000 to 1.000.
This is the cleanest case of CD providing complete rescue.

\paragraph{Instruction-semantic failures are CD-resistant.}
On \texttt{funcall\_search\_multi}, the model must emit two function calls (\texttt{search\_web} and \texttt{send\_email}).
Qwen-0.6B emits only \texttt{search\_web} across all decoders (function call score: 0.200 native $\to$ 0.200 CD).
The schema permits this (\texttt{minItems: 1}), so CD produces schema-valid but semantically incomplete output.
Token masking cannot inject task understanding.
This failure is, in principle, the target of draft-conditioned decoding \citep{reddy2026dccd}: an unconstrained draft would include both calls, and conditioning on it could preserve the intent that direct masking cannot supply.

\paragraph{The ``hollow rescue'' phenomenon.}
On the same task, Llama-1B under XGrammar scores 1.000---both calls present, all parameters present.
However, the email body reads \texttt{"1.\ 2.\ 3.\ 4.\ 5."}---list structure without content.
Structural metrics count parameter presence, not parameter quality, masking the semantic gap.

\paragraph{Returning to the motivating question.}
Section~\ref{sec:intro} asked whether \cd{} flattens the scaling curve---whether a constrained 1B model can match an unconstrained 3B model.
The answer is yes, and more: under XGrammar, Llama-1B reaches 0.986 content accuracy, \emph{exceeding} Llama-3B's native 0.777.
But the rescue is not asymmetric: the same constraint lifts Llama-3B from 0.777 to 0.991, so the constrained 1B does not match the constrained 3B---the gap narrows but does not close.
And on instruction-semantic tasks, Qwen-0.6B best-case 0.943 under XGrammar remains the lowest of all five models: the floor set by scale is lowered, not removed.
\CD{} rescues form; it does not rescue scale.

\section{Discussion}
\label{sec:discussion}

\subsection{Limitations}

Our study tests five dense models; MoE architectures under CD are left as future work.
The 14-task suite is hand-designed; evaluation on real-world API schemas would strengthen external validity.
Content accuracy uses field-level matching, which is coarse---richer semantic evaluation (e.g., LLM-as-judge) could reveal finer-grained quality differences.

\subsection{Implications for Practitioners}

For type coercion failures, XGrammar provides near-zero-overhead complete rescue.
For instruction-semantic failures, CD offers no benefit---a larger model or improved prompting is required.
The cheapest reliable configuration for structured output is XGrammar with a 3--4B model.

\section{Conclusion}
\label{sec:conclusion}

\CD{} eliminates structural failures across all tested small LLMs, but semantic correctness depends on the failure archetype and model scale.
On our motivating question: a constrained 1B model exceeds an unconstrained 3B model, yet the same constraint rescues the 3B as well---schema conformance is necessary but not sufficient for semantic correctness, and it does not remove the advantage of scale where failures are semantic.
Future work should extend to MoE architectures, real-world schemas, and richer semantic evaluation.

\bibliography{custom}

\end{document}